\documentclass[10pt,twocolumn,letterpaper]{article}

\usepackage[pagenumbers]{cvpr} 

\usepackage[dvipsnames]{xcolor}

\usepackage{amsmath,amsfonts,bm}

\def\eqref#1{equation~\ref{#1}}

\def\1{\bm{1}}

\DeclareMathAlphabet{\mathsfit}{\encodingdefault}{\sfdefault}{m}{sl}
\SetMathAlphabet{\mathsfit}{bold}{\encodingdefault}{\sfdefault}{bx}{n}

\DeclareMathOperator*{\argmin}{arg\,min}

\usepackage{graphicx}
\usepackage{amsmath}
\usepackage{amssymb}
\usepackage{booktabs}
\usepackage{enumerate}

\usepackage{amsfonts}
\usepackage{amsthm}
\usepackage{bbm}

\usepackage{wrapfig,lipsum,booktabs}
\usepackage{color}
\usepackage{algorithm,algpseudocode}
\usepackage{multicol}
\usepackage{array,multirow}

\usepackage{multirow}
\usepackage{tabularx}
\usepackage{listings}
\usepackage{xcolor}
\usepackage{color}
\usepackage{colortbl}

\makeatletter
\DeclareRobustCommand\onedot{\futurelet\@let@token\@onedot}
\def\@onedot{\ifx\@let@token.\else.\null\fi\xspace}

\def\eg{\emph{e.g}\onedot} 
\def\ie{\emph{i.e}\onedot} 
 
\def\etc{\emph{etc}\onedot}

\makeatother
\definecolor{cvprblue}{rgb}{0.21,0.49,0.74}
\usepackage[pagebackref,breaklinks,colorlinks,citecolor=cvprblue]{hyperref}

\def\paperID{11928} 
\def\confName{CVPR}
\def\confYear{2024}

\title{DragDiffusion: Harnessing Diffusion Models \\for Interactive Point-based Image Editing}

\author{Yujun Shi\textsuperscript{\rm 1}
\quad\quad
Chuhui Xue\textsuperscript{\rm 2}
\quad\quad
Jun Hao Liew\textsuperscript{\rm 2}
\quad\quad
Jiachun Pan\textsuperscript{\rm 1} \\
Hanshu Yan\textsuperscript{\rm 2}
\quad\quad
Wenqing Zhang\textsuperscript{\rm 2}
\quad\quad
Vincent Y. F. Tan\textsuperscript{\rm 1}
\quad\quad
Song Bai\textsuperscript{\rm 2} \\
\textsuperscript{\rm 1}National University of Singapore
\quad\quad \textsuperscript{\rm 2} ByteDance Inc.\\
{\tt\small shi.yujun@u.nus.edu} \quad {\tt\small vtan@nus.edu.sg} \quad
{\tt\small songbai.site@gmail.com}
}

\begin{document}

\twocolumn[{
\maketitle
\begin{center}
    \centering
    \captionsetup{type=figure}
    \vspace{-0.3cm}
    \includegraphics[width=\textwidth]{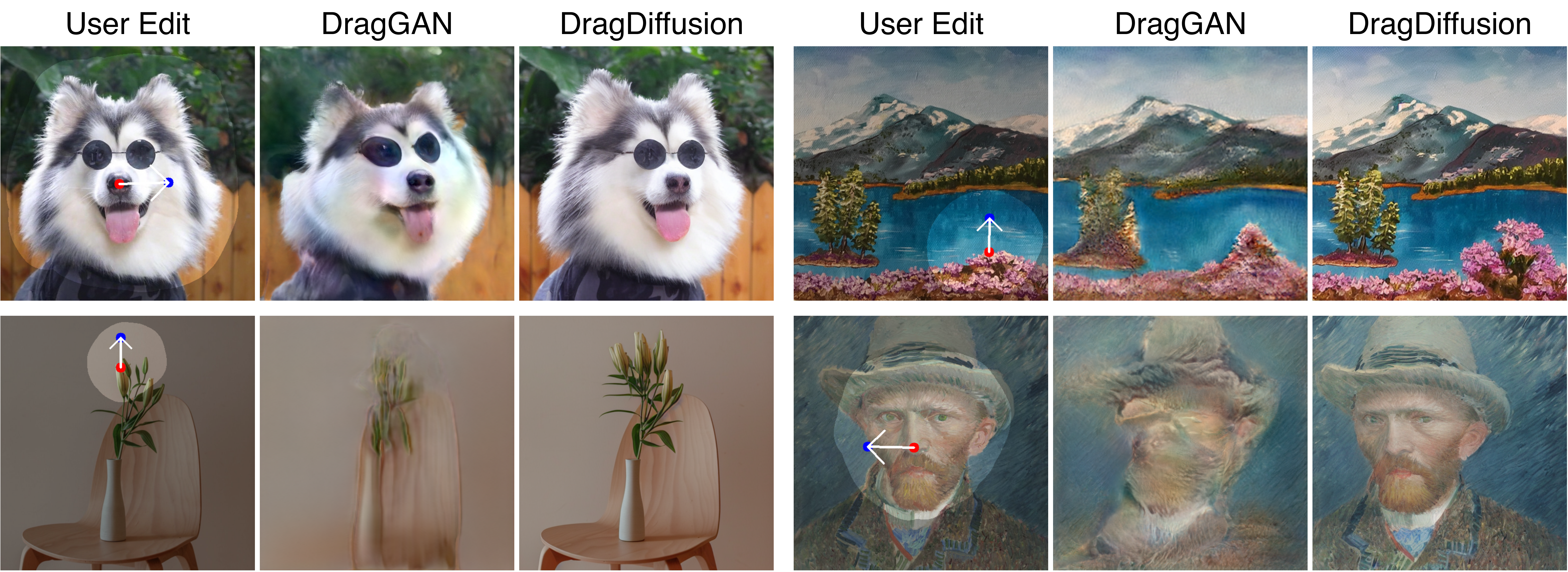}
    \captionof{figure}{\textbf{{\sc DragDiffusion} greatly improves the applicability of interactive point-based editing.} Given an input image, the user clicks handle points (\textcolor{red}{red}), target points (\textcolor{blue}{blue}), and draws a mask specifying the editable region (\textcolor{gray}{brighter area}). All results are obtained under the same user edit for fair comparisons. Project page: \href{https://yujun-shi.github.io/projects/dragdiffusion.html}{https://yujun-shi.github.io/projects/dragdiffusion.html}.
    }
    \label{fig:teaser}
\end{center}
}]

\begin{abstract}
Accurate and controllable image editing is a challenging task that has attracted significant attention recently. Notably, {\sc DragGAN} developed by Pan et al.\ (2023) \cite{pan2023drag}  is an interactive point-based image editing framework that achieves impressive editing results with pixel-level precision. However, due to its reliance on generative adversarial networks (GANs), its generality is limited by the capacity of pretrained GAN models. In this work, we extend this editing framework to diffusion models and propose a novel approach {\sc DragDiffusion}. By harnessing large-scale pretrained diffusion models, we greatly enhance the applicability of interactive point-based editing on both real and diffusion-generated images. Unlike other diffusion-based editing methods that provide guidance on diffusion latents of multiple time steps, our approach achieves efficient yet accurate spatial control by optimizing the latent of only one time step. This novel design is motivated by our observations that UNet features at a specific time step provides sufficient semantic and geometric information to support the drag-based editing. Moreover, we introduce two additional techniques, namely identity-preserving fine-tuning and reference-latent-control, to further preserve the identity of the original image. Lastly, we present a challenging benchmark dataset called {\sc DragBench}---the first benchmark to evaluate the performance of interactive point-based image editing methods. Experiments across a wide range of challenging cases (e.g., images with multiple objects, diverse object categories, various styles, etc.) demonstrate the versatility and generality of {\sc DragDiffusion}. Code and the {\sc DragBench} dataset: \href{https://github.com/Yujun-Shi/DragDiffusion}{https://github.com/Yujun-Shi/DragDiffusion}
\end{abstract}

\section{Introduction}

Image editing with generative models~\citep{roich2022pivotal,endo2022user,hertz2022prompt,mokady2023null,kawar2023imagic,parmar2023zero} has attracted extensive attention recently. One landmark work is {\sc DragGAN}~\citep{pan2023drag}, which enables interactive point-based image editing,~\ie,~drag-based editing. Under this framework, the user first clicks several pairs of handle and target points on an image. Then, the model performs semantically coherent editing on the image that moves the contents of the handle points to the corresponding target points. In addition, users can draw a mask to specify which region of the image is editable while the rest remains unchanged.

Despite {\sc DragGAN}'s impressive editing results with pixel-level spatial control, the applicability of this method is being limited by the inherent model capacity of generative adversarial networks (GANs) \citep{NIPS2014_5ca3e9b1,karras2019style,karras2020analyzing}.
On the other hand, although large-scale text-to-image diffusion models \citep{rombach2022high,saharia2022photorealistic} have demonstrated strong capabilities to synthesize high quality images across various domains, there are not many diffusion-based editing methods that can achieve precise spatial control. This is because most diffusion-based methods~\citep{hertz2022prompt,mokady2023null,kawar2023imagic,parmar2023zero} conduct editing by controlling the text embeddings, which restricts their applicability to editing high-level semantic contents or styles.

To bridge this gap, we propose {\sc DragDiffusion}, the first interactive point-based image editing method with diffusion models~\citep{rombach2022high,saharia2022photorealistic,sohl2015deep,ho2020denoising}. Empowered by large-scale pre-trained diffusion models \citep{rombach2022high,saharia2022photorealistic}, {\sc DragDiffusion} achieves accurate spatial control in image editing with significantly better generalizability (see Fig.~\ref{fig:teaser}).

\begin{figure}
    \centering
    \includegraphics[width=0.47\textwidth]{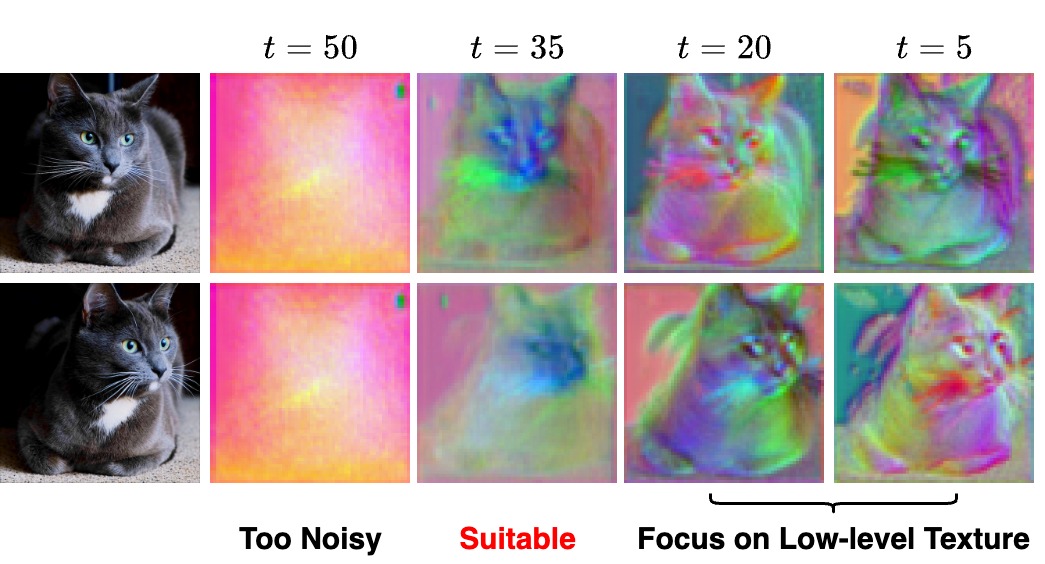}
    \caption{\textbf{PCA visualization of UNet feature maps at different diffusion steps for two video frames.}
    $t=50$ implies the full DDIM inversion, while $t=0$ implies the clean image.
    Notably, UNet features at one specific step (\eg, $t=35$) provides sufficient semantic and geometric information (\eg, shape and pose of the cat, \etc) for the drag-based editing.}
    \vspace{-.4cm}
    \label{fig:motivation_fig}
\end{figure}

\begin{figure*}
    \centering
    \includegraphics[width=0.98\textwidth]{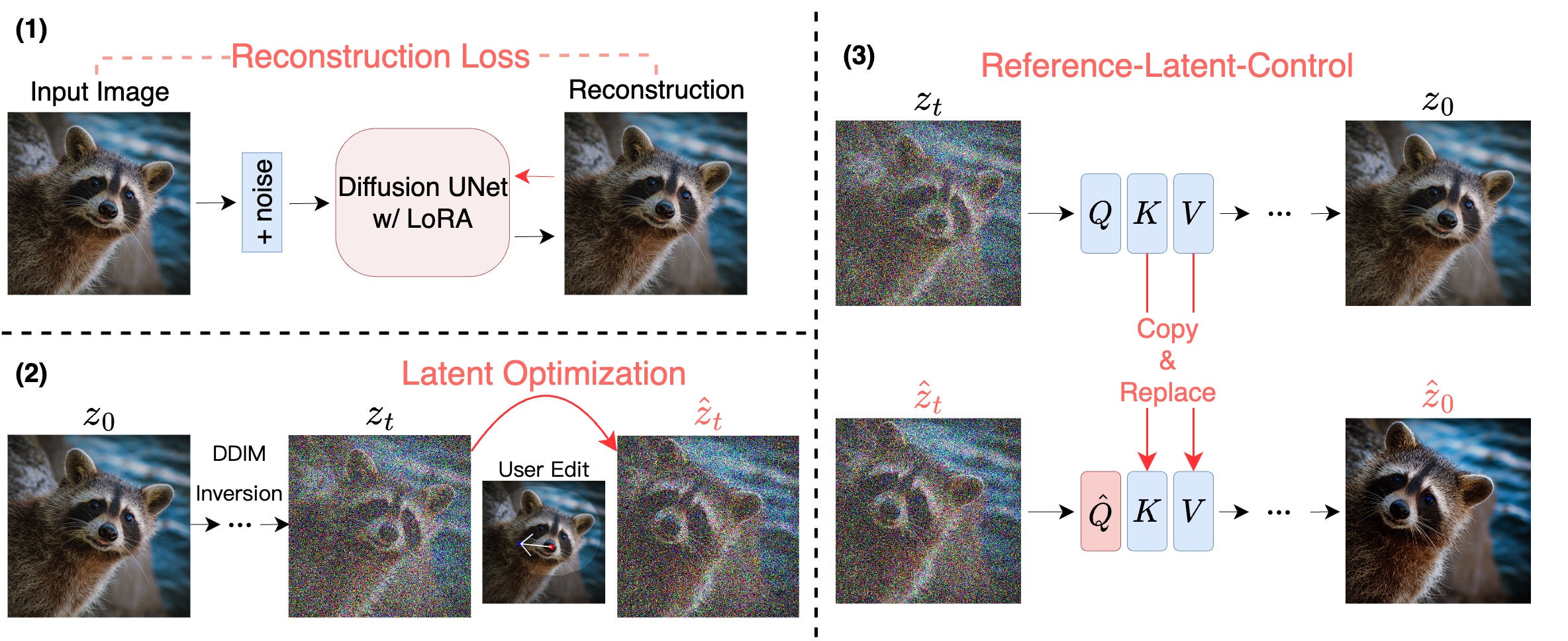}
    \vspace{-.3cm}
    \caption{\textbf{Overview of {\sc DragDiffusion}.}
    Our approach constitutes three steps: firstly, we conduct identity-preserving fine-tuning on the UNet of the diffusion model given the input image. Secondly, according to the user's dragging instruction, we optimize the latent obtained from DDIM inversion on the input image. Thirdly, we apply DDIM denoising guided by our reference-latent-control on $\hat{z}_{t}$ to obtain the final editing result $\hat{z}_{0}$. Figure best viewed in color.
    }
    \vspace{-.5cm}
    \label{fig:method_overview}
\end{figure*}

Our approach focuses on optimizing diffusion latents to achieve drag-based editing, which is inspired by the fact that diffusion latents can accurately determine the spatial layout of the generated images \cite{mao2023guided}. In contrast to previous methods \cite{parmar2023zero,epstein2023selfguidance,bansal2023universal,yu2023freedom}, which apply gradient descent on latents of {\em multiple} diffusion steps, our approach focuses on optimizing the latent of {\em one appropriately selected step} to conveniently achieve the desired editing results. This novel design is motivated by the empirical observations presented in Fig.~\ref{fig:motivation_fig}. Specifically, given two frames from a video simulating the original and the ``dragged'' images, we visualize the UNet feature maps of different diffusion steps using principal component analysis (PCA).
Via this visualization, we find that there exists a single diffusion step (\eg, $t=35$ in this case) such that the UNet feature maps at this step alone contains sufficient semantic and geometric information to support structure-oriented spatial control such as drag-based editing.
Besides optimizing the diffusion latents, we further introduce two additional techniques to enhance the identity preserving during the editing process, namely identity-preserving fine-tuning and reference-latent-control. An overview of our method is given in Fig.~\ref{fig:method_overview}.


It would be ideal to immediately evaluate our method on well-established benchmark datasets. However, due to a lack of evaluation benchmarks for interactive point-based editing, it is difficult to rigorously study and corroborate the efficacy of our proposed approach. Therefore, to facilitate such evaluation, we present {\sc DragBench}---the first benchmark dataset for drag-based editing. {\sc DragBench} is a diverse collection comprising images spanning various object categories, indoor and outdoor scenes, realistic and aesthetic styles, \emph{etc}. Each image in our dataset is accompanied with a set of ``drag'' instructions, which consists of one or more pairs of handle and target points as well as a mask specifying the editable region.

Through extensive qualitative and quantitative experiments on a variety of examples (including those on {\sc DragBench}), we demonstrate the versatility and generality of our approach. In addition, our empirical findings corroborate the crucial role played by identity-preserving fine-tuning and reference-latent-control. Furthermore, a comprehensive ablation study is conducted to meticulously explore the influence of key factors, including the number of inversion steps of the latent, the number of identity-preserving fine-tuning steps, and the UNet feature maps.

Our contributions are summarized as follows: 1) we present a novel image editing method {\sc DragDiffusion}, the first to achieve interactive point-based editing with diffusion models; 2) we introduce {\sc DragBench}, the first benchmark dataset to evaluate interactive point-based image editing methods; 3) Comprehensive qualitative and quantitative evaluation demonstrate the versatility and generality of our {\sc DragDiffusion}.

\section{Related Work}
\textbf{Generative Image Editing.} Given the initial successes of generative adversarial networks (GANs) in image generation \citep{NIPS2014_5ca3e9b1,karras2019style,karras2020analyzing}, many previous image editing methods have been based on the GAN paradigm \citep{endo2022user,pan2023drag,abdal2021styleflow,leimkuhler2021freestylegan,patashnik2021styleclip,shen2020interpreting,shen2021closed,tewari2020stylerig,harkonen2020ganspace,zhu2016generative,zhu2023linkgan}. However, due to the limited model capacity of GANs and the difficulty of inverting the real images into GAN latents \citep{abdal2019image2stylegan,creswell2018inverting,lipton2017precise,roich2022pivotal}, the generality of these methods would inevitably be constrained. Recently, due to the impressive generation results from large-scale text-to-image diffusion models \citep{rombach2022high,saharia2022photorealistic}, many diffusion-based image editing methods have been proposed \citep{hertz2022prompt,cao2023masactrl,mao2023guided,kawar2023imagic,parmar2023zero,liew2022magicmix,mou2023dragondiffusion,tumanyan2023plug,brooks2023instructpix2pix,meng2021sdedit,bar2022text2live}. Most of these methods aim to edit the images by manipulating the prompts of the image. However, as many editing attempts are difficult to convey through text, the prompt-based paradigm usually alters the image's high-level semantics or styles, lacking the capability of achieving precise pixel-level spatial control. \cite{epstein2023selfguidance} is one of the early efforts in exploring better controllability on diffusion models beyond the prompt-based image editing. In our work, we aim at enabling a even more versatile paradigm than the one studied in \cite{epstein2023selfguidance} with diffusion models---interactive point-based image editing. 

\textbf{Point-based editing.} To enable fine-grained editing, several works have been proposed to perform point-based editing, such as \cite{pan2023drag,endo2022user,wang2022rewriting}. In particular, {\sc DragGAN} has demonstrated impressive dragging-based manipulation with two simple ingredients: 1) optimization of latent codes to move the handle points towards their target locations and 2) a point tracking mechanism that keep tracks of the handle points. However, its generality is constrained due to the limited capacity of GAN. FreeDrag \citep{ling2023freedrag} propose to improve {\sc DragGAN} by introducing a point-tracking-free paradigm. In this work, we extend the editing framework of {\sc DragGAN} to diffusion models and showcase its generality over different domains. 
There is a work \citep{mou2023dragondiffusion} concurrent to ours that also studies drag-based editing with diffusion models. Differently, they rely on classifier guidance to transforms the editing signal into gradients.

\textbf{LoRA in Diffusion Models.}
Low Rank Adaptation (\ie, LoRA) \citep{hu2021lora} is a general technique to conduct parameter-efficient fine-tuning on large and deep networks. During LoRA fine-tuning, the original weights of the model are frozen, while trainable rank decomposition matrices are injected into each layer. The core assumption of this strategy is that the model weights will primarily be adapted within a low rank subspace during fine-tuning. While LoRA was initially introduced for adapting language models to downstream tasks, recent efforts have illustrated its effectiveness when applied in conjunction with diffusion models \citep{lora_stable, gu2023mix}. In this work, inspired by the promising results of using LoRA for image generation and editing \citep{ruiz2023dreambooth,kawar2023imagic}, we also implement our identity-preserving fine-tuning with LoRA.

\section{Methodology}
\label{sec:method}
In this section, we formally present the proposed {\sc DragDiffusion} approach. To commence, we introduce the preliminaries on diffusion models. Then, we elaborate on the three key stages of our approach as depicted in Fig.~\ref{fig:method_overview}: 1) identity-preserving fine-tuning; 2) latent optimization according to the user-provided dragging instructions; 3) denoising the optimized latents guided by our reference-latent-control.

\subsection{Preliminaries on Diffusion Models}
\label{sec:preliminaries}
Denoising diffusion probabilistic models (DDPM) \citep{sohl2015deep,ho2020denoising} constitute a family of latent generative models. Concerning a data distribution $q(z)$, DDPM approximates $q(z)$ as the marginal $p_{\theta}(z_0)$ of the joint distribution between $Z_{0}$ and a collection of latent random variables $Z_{1:T}$. Specifically,
\begin{equation}
    p_{\theta}(z_{0}) = \int{p_{\theta}(z_{0:T})\, \mathrm{d}z_{1:T}},
\end{equation}
where $p_{\theta}(z_T)$ is a standard normal distribution and the transition kernels $p_{\theta}(z_{t-1}|z_t)$ of this Markov chain are all Gaussian conditioned on $z_t$.
In our context, $Z_{0}$ corresponds to image samples given by users, and $Z_{t}$ corresponds to the latent after $t$ steps of the diffusion process.

\cite{rombach2022high} proposed the latent diffusion model (LDM), which maps data into a lower-dimensional space via a variational auto-encoder (VAE) \citep{kingma2013auto} and models the distribution of the latent embeddings instead. Based on the framework of LDM, several powerful pretrained diffusion models have been released publicly, including the Stable Diffusion (SD) model (\href{https://huggingface.co/stabilityai}{https://huggingface.co/stabilityai}). In SD, the network responsible for modeling $p_{\theta}(z_{t-1}|z_t)$ is implemented as a UNet \citep{ronneberger2015u} that comprises multiple self-attention and cross-attention modules \citep{vaswani2017attention}. Applications in this paper are implemented based on the public Stable Diffusion model.

\subsection{Identity-preserving Fine-tuning}
\label{sec:lora_finetune}
Before editing a real image, we first conduct identity-preserving fine-tuning \citep{hu2021lora} on the diffusion models' UNet (see  panel~(1) of Fig.~\ref{fig:method_overview}). This stage aims to ensure that the diffusion UNet encodes the features of input image more accurately (than in the absence of this procedure), thus facilitating the  consistency of the identity of the image throughout the editing process.
This fine-tuning process is implemented with LoRA \cite{hu2021lora}. More formally, the objective function of the LoRA is
\begin{equation}
    \mathcal{L}_{\text{ft}}(z, \Delta\theta) =  \mathbb{E}_{\epsilon, t} \left[\|\epsilon - \epsilon_{\theta + \Delta\theta}(\alpha_{t}z + \sigma_{t}\epsilon)\|^{2}_{2}\right],
\end{equation}
where $\theta$ and $\Delta\theta$ represent the UNet and LoRA parameters respectively, $z$ is the real image, $\epsilon \sim \mathcal{N}(\mathbf{0},\mathbf{I})$ is the randomly sampled noise map, $\epsilon_{\theta+\Delta\theta}(\cdot)$ is the noise map predicted by the LoRA-integrated UNet, and $\alpha_t$ and $\sigma_t$ are parameters of the diffusion noise schedule at diffusion step $t$. The fine-tuning objective is optimized via gradient descent on $\Delta\theta$.

Remarkably, we find that fine-tuning LoRA for merely $80$ steps proves sufficient for our approach, which is in stark contrast to the $1000$ steps required by tasks such as subject-driven image generation \cite{ruiz2023dreambooth,gu2023mix}. This ensures that our identity-preserving fine-tuning process is extremely efficient, and only takes around $25$ seconds to complete on an A100 GPU. We posit this efficiency is because our approach operates on the inverted noisy latent, which inherently preserve some information about the input real image. Consequently, our approach does not require lengthy fine-tuning to preserve the identity of the original image.

\subsection{Diffusion Latent Optimization}
\label{sec:latent_optimization}
After identity-preserving fine-tuning, we optimize the diffusion latent according to the user instruction (\ie, the handle and target points, and optionally a mask specifying the editable region) to achieve the desired interactive point-based editing (see panel~(2) of Fig.~\ref{fig:method_overview}).

To commence, we first apply a DDIM inversion \citep{song2020denoising} on the given real image to obtain a diffusion latent at a certain step $t$ (\ie, $z_t$). This diffusion latent serves as the initial value for our latent optimization process. Then, following along the similar spirit of \cite{pan2023drag}, the latent optimization process consists of two steps to be implemented consecutively. These two steps, namely motion supervision and point tracking, are executed repeatedly until either all handle points have moved to the targets or the maximum number of iterations has been reached. Next, we describe these two steps in detail.

\textbf{Motion Supervision:} We denote the $n$ handle points at the $k$-th motion supervision iteration as $\{h^{k}_{i}=(x^{k}_{i},y^{k}_{i}): i=1,\ldots,n\}$ and their corresponding target points as $\{g_{i}=(\tilde{x}_{i},\tilde{y}_{i}):i=1,\ldots,n\}$. The input image is denoted as $z_{0}$; the $t$-th step latent (\ie, result of $t$-th step DDIM inversion) is denoted as $z_{t}$.
We denote the UNet output feature maps used for motion supervision as $F(z_{t})$, and the feature vector at pixel location $h^{k}_{i}$ as $F_{h^{k}_{i}}(z_{t})$.
Also, we denote the square patch centered around $h^{k}_{i}$ as $\Omega(h^{k}_{i}, r_{1})=\{(x,y) : |x-x^{k}_{i}| \leq r_{1}, |y-y^{k}_{i}| \leq r_{1}\}$.
Then, the motion supervision loss at the $k$-th iteration is defined as:
\begin{align}
    \!\mathcal{L}_{\text{ms}}(\hat{z}^{k}_{t}) &= \sum_{i=1}^{n}\sum_{q\in\Omega(h^{k}_{i}, r_{1})} \left\|F_{q+d_{i}}(\hat{z}^{k}_{t})-\mathrm{sg}(F_{q}(\hat{z}^{k}_{t})) \right\|_{1} 
    \nonumber\\
    &\quad+\lambda\left\|(\hat{z}^{k}_{t-1} - \mathrm{sg}(\hat{z}^{0}_{t-1}))\odot(\mathbbm{1} \!-\!M)\right\|_{1},\label{eq: sup_obj}
\end{align}
where $\hat{z}^{k}_{t}$ is the $t$-th step latent after the $k$-th update, $\mathrm{sg} (\cdot)$ is the stop gradient operator (\ie, the gradient will not be back-propagated for the term $\mathrm{sg}(F_{q}(\hat{z}_{t}^{k}))$), $d_{i} = (g_i - h^{k}_i)/\|g_i - h^{k}_i\|_{2}$ is the normalized vector pointing from $h^{k}_i$ to $g_i$, $M$ is the binary mask specified by the user, $F_{q+d_{i}}(\hat{z}_{t}^{k})$ is obtained via bilinear interpolation as the elements of $q+d_{i}$ may not be integers. In each iteration, $\hat{z}^{k}_{t}$ is updated by taking one gradient descent step to minimize $\mathcal{L}_{\text{ms}}$:
\begin{equation}
    \hat{z}^{k+1}_{t} = \hat{z}^{k}_{t} - \eta\cdot\frac{\partial \mathcal{L}_{\text{ms}}(\hat{z}^{k}_{t})}{\partial \hat{z}^{k}_{t}},
\end{equation}
where $\eta$ is the learning rate for latent optimization.

Note that for the second term in Eqn.~(\ref{eq: sup_obj}), which encourages the unmasked area to remain unchanged, we are working with the diffusion latent instead of the UNet features. Specifically, given $\hat{z}_{t}^{k}$, we first apply one step of DDIM denoising to obtain $\hat{z}_{t-1}^{k}$, then we regularize the unmasked region of $\hat{z}_{t-1}^{k}$ to be the same as $\hat{z}_{t-1}^{0}$ (\ie, $z_{t-1}$).

\textbf{Point Tracking:} Since the motion supervision updates $\hat{z}^{k}_{t}$, the positions of the handle points may also change. Therefore, we need to perform point tracking to update the handle points after each motion supervision step. To achieve this goal, we use UNet feature maps $F(\hat{z}_{t}^{k+1})$ and $F(z_{t})$ to track the new handle points. Specifically, we update each of the handle points $h_i^k$ with a nearest neighbor search within the square patch $\Omega(h^{k}_{i}, r_{2}) = \{(x,y) : |x-x^{k}_{i}| \leq r_{2}, |y-y^{k}_{i}| \leq r_{2}\}$ as follows:
\begin{equation}
\label{eq: point_track}
    h_{i}^{k+1} =\argmin_{q\in\Omega(h^{k}_{i},r_{2})} \left\|F_{q}(\hat{z}^{k+1}_{t}) - F_{h^{0}_{i}}(z_{t}) \right\|_{1}.
\end{equation}

\subsection{Reference-latent-control}
\label{sec:latent_masactrl}
After we have completed the optimization of the diffusion latents, we then denoise the optimized latents to obtain the final editing results. However, we find that na\"ively applying DDIM denoising on the optimized latents still occasionally leads to undesired identity shift or degradation in quality comparing to the original image. We posit that this issue arises due to the absence of adequate guidance from the original image during the denoising process.


To mitigate this problem, we draw inspiration from \citep{cao2023masactrl} and propose to leverage the property of self-attention modules to steer the denoising process, thereby boosting coherence between the original image and the editing results.
In particular, as illustrated in panel~(3) of Fig.~\ref{fig:method_overview}, given the denoising process of both the original latent $z_{t}$ and the optimized latent $\hat{z}_{t}$, we use the process of $z_{t}$ to guide the process of $\hat{z}_{t}$. More specifically, during the forward propagation of the UNet's self-attention modules in the denoising process, we replace the key and value vectors generated from $\hat{z}_{t}$ with the ones generated from $z_{t}$. With this simple replacement technique, the query vectors generated from $\hat{z}_{t}$ will be directed to query the correlated contents and texture of $z_{t}$. This leads to the denoising results of $\hat{z}_{t}$ (\ie, $\hat{z}_{0}$) being more coherent with the denoising results of $z_t$ (\ie, $z_0$). In this way, reference-latent-control substantially improves the consistency between the original and the edited images.

\section{Experiments}
\subsection{Implementation Details}
\label{sec:impl_details}
In all our experiments, unless stated otherwise, we adopt the Stable Diffusion 1.5 \citep{rombach2022high} as our diffusion model. During the identity-preserving fine-tuning, we inject LoRA into the projection matrices of query, key and value in all of the attention modules.
We set the rank of the LoRA to $16$. We fine-tune the LoRA using the AdamW~\citep{kingma2014adam} optimizer with a learning rate of $5\times 10^{-4}$ and a batch size of $4$ for $80$ steps.

During the latent optimization stage, we schedule $50$ steps for DDIM and optimize the diffusion latent at the $35$-th step unless specified otherwise. When editing real images, we {\em do not} apply classifier-free guidance (CFG)~\citep{ho2022classifier} in both DDIM inversion and DDIM denoising process. This is because CFG tends to amplify numerical errors, which is not ideal in performing the DDIM inversion \citep{mokady2023null}. We use the Adam optimizer with a learning rate of $0.01$ to optimize the latent. The maximum optimization step is set to be $80$. The hyperparameter $r_{1}$ in Eqn.~\ref{eq: sup_obj} and $r_{2}$ in Eqn.~\ref{eq: point_track} are tuned to be $1$ and $3$, respectively. $\lambda$ in Eqn.~\ref{eq: sup_obj} is set to $0.1$ by default, but the user may increase $\lambda$ if the unmasked region has changed to be more than what was desired.

Finally, we apply our reference-latent-control in the upsampling blocks of the diffusion UNet at all denoising steps when generating the editing results. The execution time for each component is detailed in Appendix~D.

\subsection{{\sc DragBench} and Evaluation Metrics}
\label{sec:dragbench_evaluation}
Since interactive point-based image editing is a recently introduced paradigm, there is an absence of dedicated evaluation benchmarks for this task, making it challenging to comprehensively study the effectiveness of our proposed approach. To address the need for systematic evaluation, we introduce {\sc DragBench}, the first benchmark dataset tailored for drag-based editing. {\sc DragBench} is a diverse compilation encompassing various types of images. Details and examples of our dataset are given in Appendix~A. Each image within our dataset is accompanied by a set of dragging instructions, comprising one or more pairs of handle and target points, along with a mask indicating the editable region. We hope future research on this task can benefit from {\sc DragBench}.

In this work, we utilize the following two metrics for quantitative evaluation: {\em  Image Fidelity} (IF) \citep{kawar2023imagic} and {\em  Mean Distance} (MD) \citep{pan2023drag}. IF, the first metric, quantifies the similarity between the original and edited images.
It is calculated by subtracting the mean LPIPS \citep{zhang2018unreasonable} over all pairs of original and edited images from $1$.
The second metric MD assesses how well the approach moves the semantic contents to the target points. To compute the MD, we first employ DIFT \citep{tang2023emergent} to identify points in the edited images corresponding to the handle points in the original image. These identified points are considered to be the final handle points post-editing. MD is subsequently computed as the mean Euclidean distance between positions of all target points and their corresponding final handle points.
MD is averaged over all pairs of handle and target points in the dataset.
An optimal ``drag'' approach ideally achieves both a low MD (indicating effective ``dragging'') and a high IF (reflecting robust identity preservation).

\begin{figure*}
    \centering
    \includegraphics[width=\textwidth]{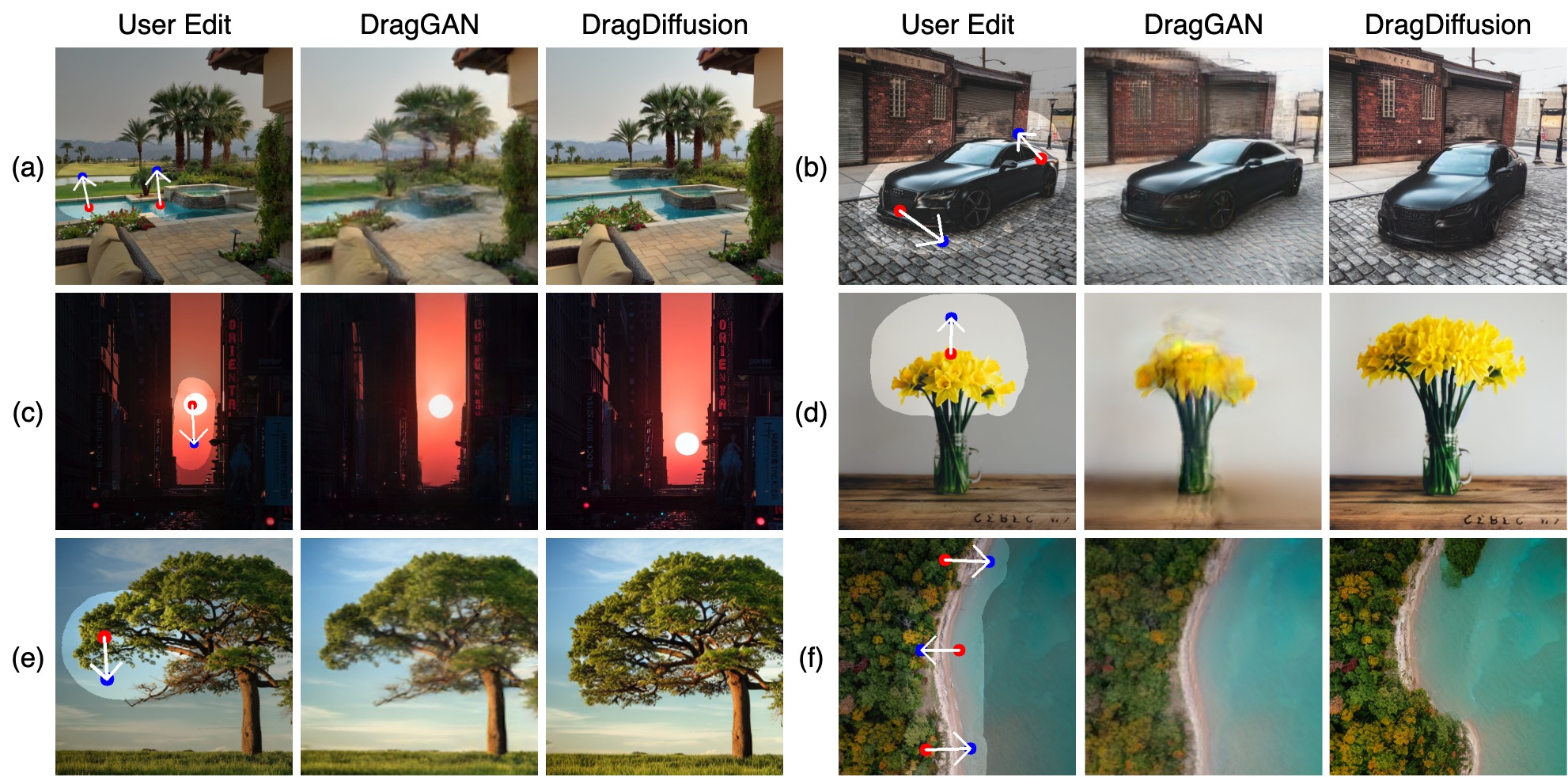}
    \vspace{-.4cm}
    \caption{Comparisons between {\sc DragGAN} and {\sc DragDiffusion}.
    All results are obtained under the same user edit for fair comparisons.
    }
    \vspace{-.3cm}
    \label{fig:compare_draggan}
\end{figure*}

\begin{figure*}
    \centering
    \includegraphics[width=\textwidth]{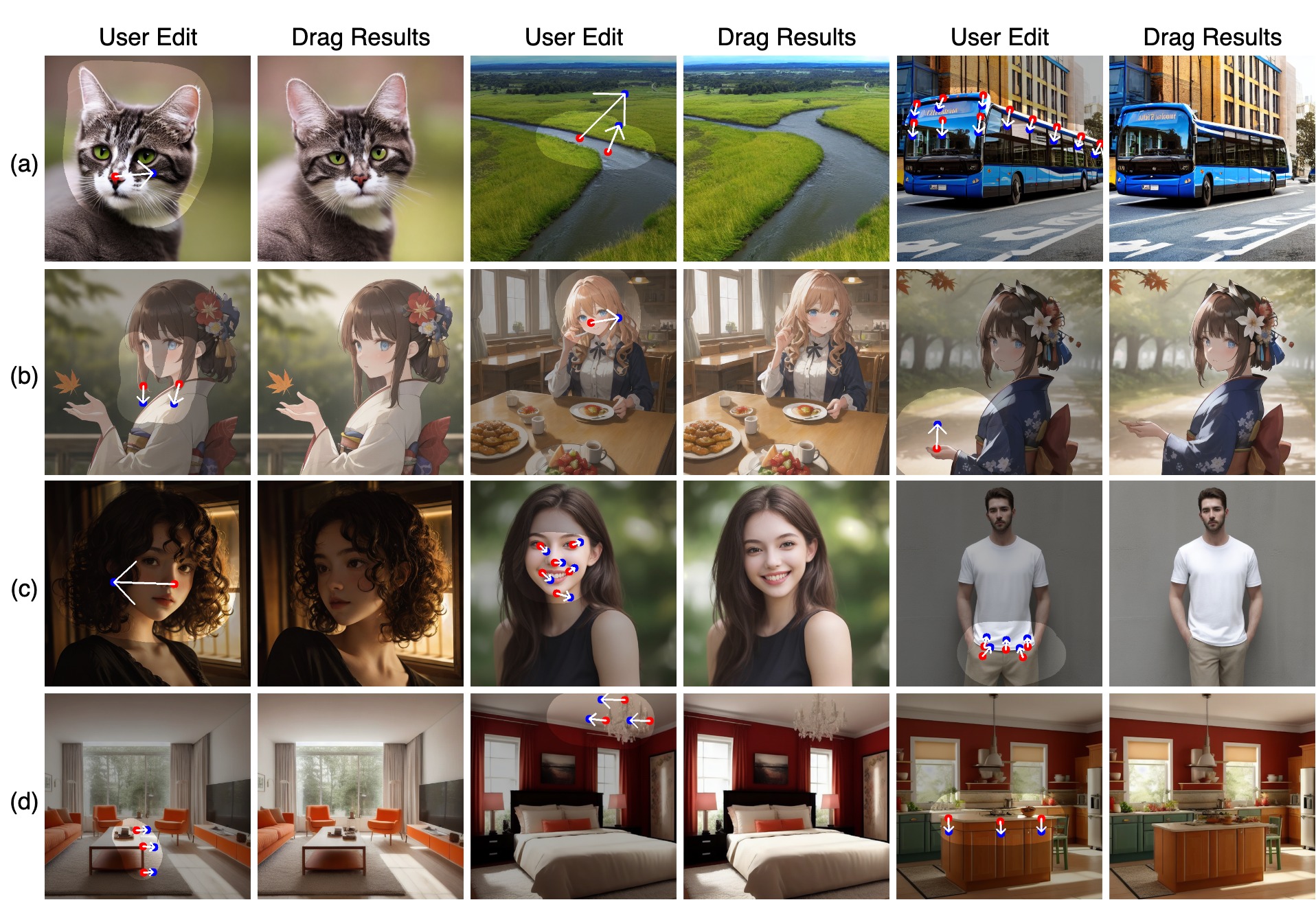}
    \vspace{-.4cm}
    \caption{Editing results on diffusion-generated images with \textbf{(a)} Stable-Diffusion-1.5, \textbf{(b)} Counterfeit-V2.5,
    \textbf{(c)} Majicmix Realistic,
    \textbf{(d)} Interior Design Supermix.}
    \label{fig:quality_eval}
\end{figure*}

\begin{figure*}
    \centering
    \includegraphics[width=\textwidth]{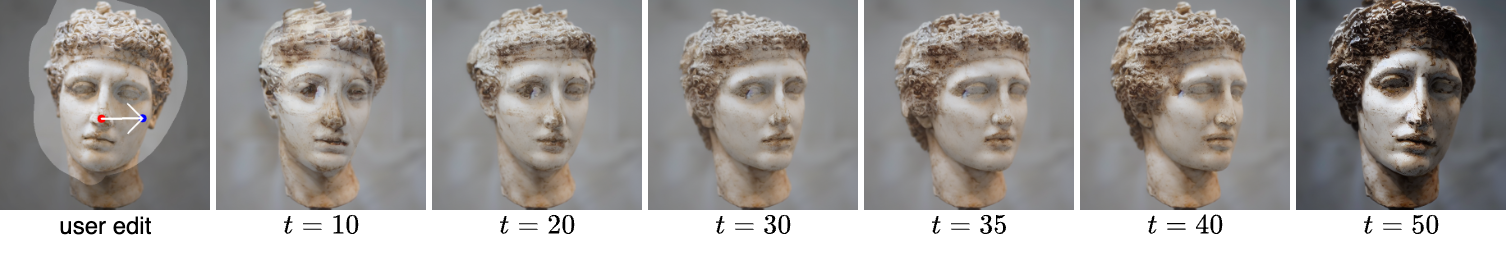}
    \vspace{-.8cm}
    \caption{Ablating the number of inversion step $t$. Effective results are obtained when $t\in [30,40]$.}
    \vspace{-.3cm}
    \label{fig:qualitative_ablate_inv}
\end{figure*}

\subsection{Qualitative Evaluation}
In this section, we first compare our approach with {\sc DragGAN} on real images. We employ SD-1.5 for our approach when editing real images. All input images and the user edit instructions are from our {\sc DragBench} dataset. Results are given in Fig.~\ref{fig:compare_draggan}. As illustrated in the figure, when dealing with the real images from a variety of domains, {\sc DragGAN} often struggles due to GAN models' limited capacity. On the other hand, our {\sc DragDiffusion} can convincingly deliver reasonable editing results.
More importantly, besides achieving the similar pose manipulation and local deformation as in {\sc DragGAN} \cite{pan2023drag}, our approach even enables more types of editing such as content filling. An example is given in Fig.~\ref{fig:compare_draggan}~(a), where we fill the grassland with the pool using drag-based editing. This further validates the enhanced versatility of our approach. More qualitative comparisons are provided in Appendix~F.

Next, to show the generality of our approach, we perform drag-based editing on diffusion-generated images across a spectrum of variants of SD-1.5, including SD-1.5 itself, Counterfeit-V2.5, Majicmix Realistic, Interior Design Supermix. Results are shown in Fig.~\ref{fig:quality_eval} These results validate our approach's ability to smoothly work with various pre-trained diffusion models.
Moreover, these results also illustrate our approach's ability to deal with drag instructions of different magnitudes (\eg, small magnitude edits such as the left-most image in Fig.~\ref{fig:quality_eval}~(d) and large magnitude edits such as the left-most image in Fig.~\ref{fig:quality_eval}~(c)).
Additional results with more diffusion models and different resolutions can be found in Appendix~F.

\begin{figure*}
    \centering
    \includegraphics[width=\textwidth]{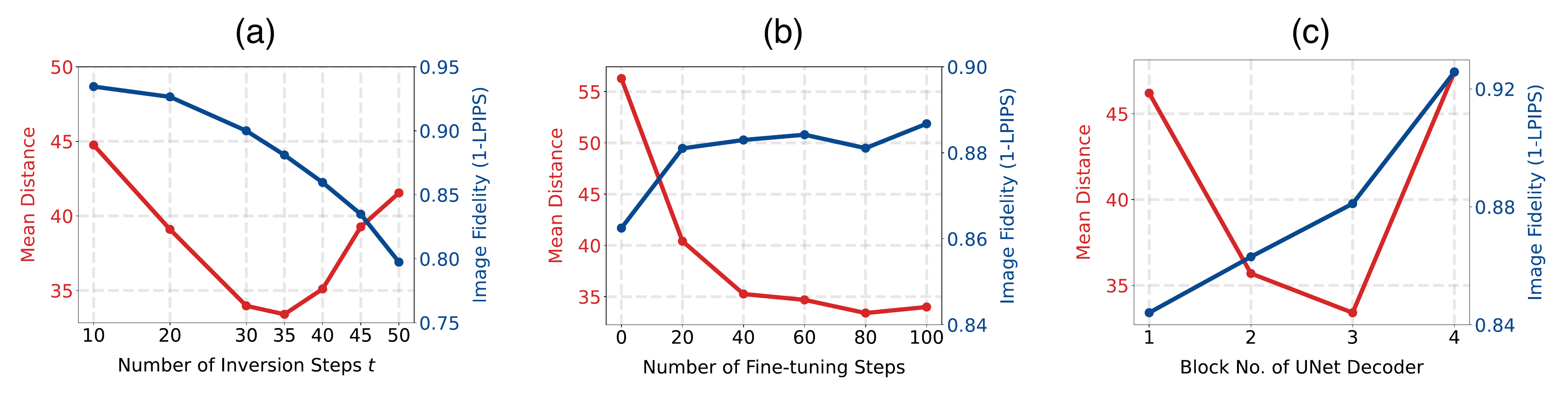}
    \vspace{-0.6cm}
    \caption{Ablation study on \textbf{(a)} the number of inversion step $t$ of the diffusion latent; \textbf{(b)} the number of identity-preserving fine-tuning steps; \textbf{(c)} Block No.\ of UNet feature maps. Mean Distance ($\downarrow$) and Image Fidelity ($\uparrow$) are reported. Results are produced on {\sc DragBench}.}
    \vspace{-0.2cm}
    \label{fig:ablation}
\end{figure*}

\begin{figure}
    \centering
    \includegraphics[width=0.47\textwidth]{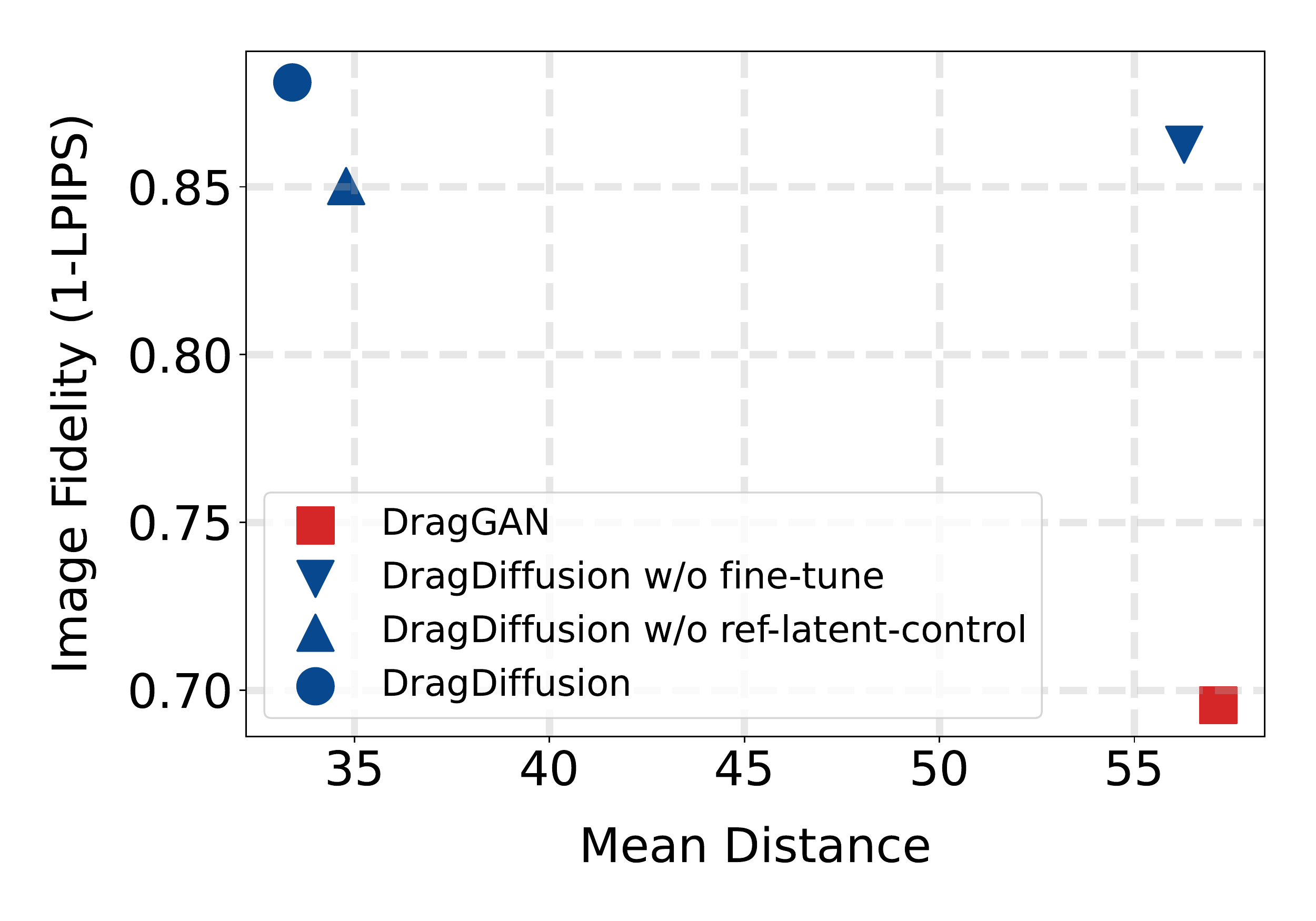}
    \vspace{-0.4cm}
    \caption{Quantitative analysis on {\sc DragGAN}, {\sc DragDiffusion} and {\sc DragDiffusion}'s variants without certain components. Image Fidelity ($\uparrow$) and Mean Distance ($\downarrow$) are reported. Results are produced on {\sc DragBench}. The approach with better results should locate at the upper-left corner of the coordinate plane.}
    \vspace{-0.1cm}
    \label{fig:quant_compare}
\end{figure}

\begin{figure}
    \centering
    \includegraphics[width=0.47\textwidth]{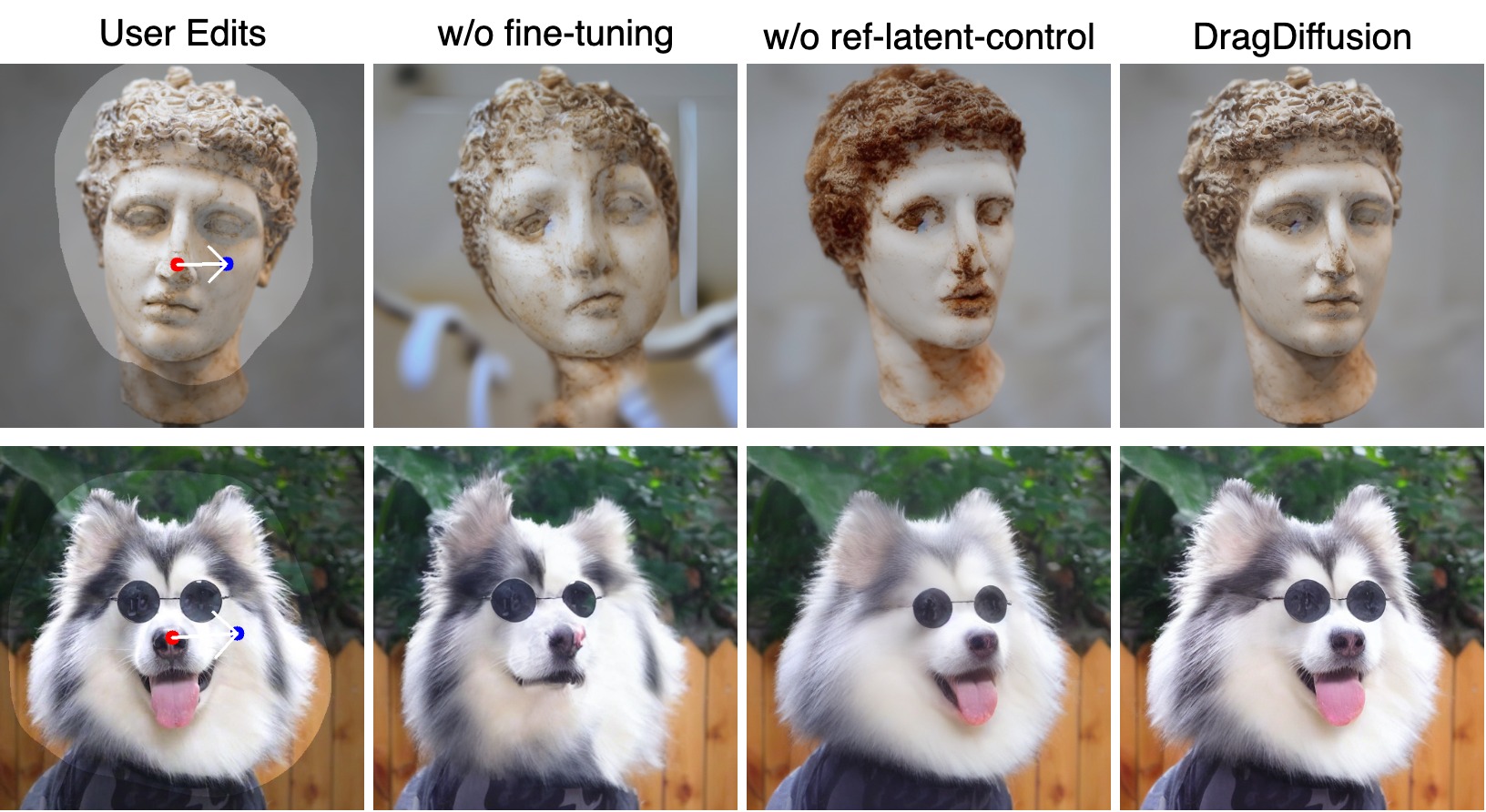}
    \vspace{-0.2cm}
    \caption{Qualitative validation on effectiveness of identity-preserving fine-tuning and reference-latent-control.}
    \label{fig:qualitative_ft_rlc}
    \vspace{-0.4cm}
\end{figure}

\subsection{Quantitative Analysis}
In this section, we conduct a rigorous quantitative evaluation to assess the performance of our approach. We begin by comparing   {\sc DragDiffusion} with the baseline method {\sc DragGAN}. As each StyleGAN \citep{karras2020analyzing} model utilized in \cite{pan2023drag} is specifically designed for a particular image class, we employ an ensemble strategy to evaluate {\sc DragGAN}. This strategy involves assigning a text description to characterize the images generated by each StyleGAN model. Before editing each image, we compute the CLIP similarity \citep{radford2021learning} between the image and each of the text descriptions associated with the GAN models. The GAN model that yields the highest CLIP similarity is selected for the editing task.

Furthermore, to validate the effectiveness of each component of our approach, we evaluate {\sc DragDiffusion} in the following two configurations: one without identity-preserving fine-tuning and the other without reference-latent-control. We perform our empirical studies on the {\sc DragBench} dataset, and Image Fidelity (IF) and Mean Distance (MD) of each configuration mentioned above are reported in Fig.~\ref{fig:quant_compare}. All results are averaged over the {\sc DragBench} dataset. In this figure, the $x$-axis represents MD and the $y$-axis represents IF, which indicates the method with better results should locate at the upper-left corner of the coordinate plane. The results presented in this figure clearly demonstrate that our {\sc DragDiffusion} significantly outperforms the {\sc DragGAN} baseline in terms of both IF and MD.
Furthermore, we observe that {\sc DragDiffusion} without identity-preserving fine-tuning experiences a catastrophic increase in MD, whereas {\sc DragDiffusion} without reference-latent control primarily encounters a decrease in IF.
Visualization on the effects of identity-preserving fine-tuning and reference-latent-control are given in Fig.~\ref{fig:qualitative_ft_rlc}, which corroborates with our quantitative results.


\subsection{Ablation on the Number of Inversion Step}
Next, we conduct an ablation study to elucidate the impact of varying $t$ (\ie, the number of inversion steps) during the latent optimization stage of {\sc DragDiffusion}. We set $t$ to be $t=10,20,30,40,50$ steps and run our approach on {\sc DragBench} to obtain the editing results ($t=50$ corresponds to the pure noisy latent). We evaluate Image Fidelity (IF) and Mean Distance (MD) for each $t$ value in Fig.~\ref{fig:ablation}(a). All metrics are averaged over the {\sc DragBench} dataset.

In terms of the IF, we observe a monotonic decrease as $t$ increases. This trend can be attributed to the stronger flexibility of the diffusion latent as more steps are inverted. As for MD, it initially decreases and then increases with higher $t$ values. This behavior highlights the presence of a critical range of $t$ values for effective editing ($t\in [30,40]$ in our figure). When $t$ is too small, the diffusion latent lacks the necessary flexibility for substantial changes, posing challenges in performing reasonable edits. Conversely, overly large $t$ values result in a diffusion latent that is unstable for editing, leading to difficulties in preserving the original image's identity. Given these results, we chose $t=35$ as our default setting, as it achieves the lowest MD while maintaining a decent IF. Qualitative visualization that corroborates with our numerical evaluation is provided in Fig.~\ref{fig:qualitative_ablate_inv}.

\subsection{Ablation Study on the Number of Identity-preserving Fine-tuning Steps}
We run our approach on {\sc DragBench} under $0$, $20$, $40$, $60$, $80$, and $100$ identity-preserving fine-tuning steps, respectively ($0$ being no fine-tuning). The outcomes are assessed using IF and MD, and the results are presented in Fig.\ref{fig:ablation}~(b). All results are averaged over the {\sc DragBench} dataset.

Initially, as the number of fine-tuning steps increases, MD exhibits a steep downward trend while IF shows an upward trend. This reflects that identity-preserving fine-tuning can drastically boost both the precision and consistency of drag-based editing. However, as the fine-tuning progresses, both MD and IF subsequently begins to plateau. This phenomenon shows that lengthy fine-tuning of LoRA would no longer significantly improve the performance of our approach. Considering the experimental results, we conduct identity-preserving fine-tuning for $80$ steps by default to balance between effectiveness and efficiency. Visualizations that corroborate our quantitative evaluation are presented in the Appendix~G.


\subsection{Ablation Study on the UNet Feature Maps}
Finally, we study the effect of using different blocks of UNet feature maps to supervise our latent optimization. We run our approach on the {\sc DragBench} dataset with the feature maps output by $4$ different upsampling blocks of the UNet \textit{Decoder}, respectively. The outcomes are assessed with IF and MD, and are shown in Fig.~\ref{fig:ablation}(c). As can be seen, with deeper blocks of UNet features, IF consistently increases, while MD first decreases and then increases. This trend is because feature maps of lower blocks contain coarser semantic information, while higher blocks contain lower level texture information \cite{tumanyan2023plug,tokenflow2023}. Hence, the feature maps of lower blocks (\eg, block No.\ of $1$) lack fine-grained information for accurate spatial control, whereas those of higher blocks (\eg, block No.\ of $4$) lack semantic and geometric information to drive the drag-based editing. Our results indicate that the feature maps produced by the third block of the UNet decoder demonstrate the best performance, exhibiting the lowest MD and a relatively high IF. Visualizations that corroborate  our quantitative evaluation are presented in the Appendix~H.


\section{Conclusion and Future Works}
In this work, we extend interactive point-based editing to large-scale pretrained diffusion models through the introduction of a novel method named {\sc DragDiffusion}. Furthermore, we introduce the {\sc DragBench} dataset, which aims to facilitate the evaluation of the interactive point-based editing methods. Comprehensive qualitative and quantitative results showcases the remarkable versatility and generality of our proposed method. Limitations of our approach are further discussed in Appendix~E, and we leave making the drag-based editing more robust and reliable on diffusion models as our future work.

{
    \small
    \bibliographystyle{ieeenat_fullname}
    \bibliography{main}
}

\clearpage

\onecolumn
\appendix
\maketitle
\thispagestyle{empty}

\section{Details About {\sc DragBench} Dataset}
\label{app:dragbench}
\begin{figure}
    \centering
    \includegraphics[width=\textwidth]{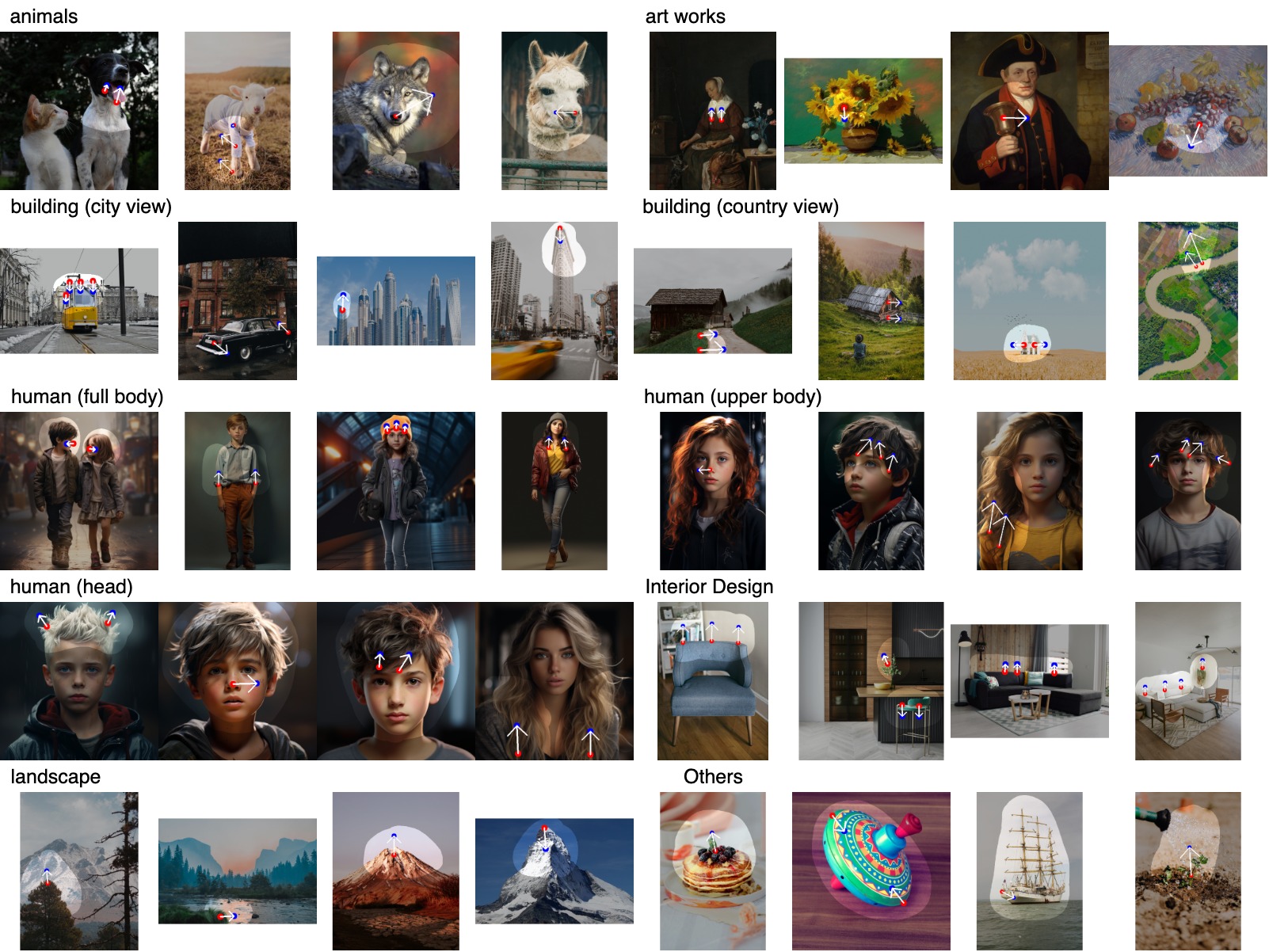}
    \caption{Examples of our {\sc DragBench} dataset. Each image is accompanied by a set of drag-based editing instruction.}
    \label{app:dragbench_example}
\end{figure}

We have collected $205$ images and provided $349$ pairs of handle and target points in total. Images in our {\sc DragBench} are classified into the following $10$ categories: animals, art works, buildings (city view), buildings (countryside view), human (head), human (upper body), human (full body), interior design, landscape, other objects. All human-related images are selected from Midjourney generation results to avoid potential legal concerns. All the other images are real images downloaded from unsplash (\href{https://unsplash.com/}{https://unsplash.com/}), pexels (\href{https://www.pexels.com/zh-cn/}{https://www.pexels.com/zh-cn/}), and pixabay (\href{https://pixabay.com/}{https://pixabay.com/}). Some examples of our dataset is given in Fig.~\ref{app:dragbench_example}.

\section{Links to the Stable Diffusion's Finetuned Variants Used by Us}
\label{app:link_to_models}
Here, we provide links to the fine-tuned variants of Stable Diffusion used by us:

Counterfeit-V2.5 (\href{https://huggingface.co/gsdf/Counterfeit-V2.5}{https://huggingface.co/gsdf/Counterfeit-V2.5}), 

Majixmix Realistic (\href{https://huggingface.co/emilianJR/majicMIX_realistic}{https://huggingface.co/emilianJR/majicMIX\_realistic}),

Realistic Vision (\href{https://huggingface.co/SG161222/Realistic_Vision_V2.0}{https://huggingface.co/SG161222/Realistic\_Vision\_V2.0}),

Interior Design Supermix (\href{https://huggingface.co/stablediffusionapi/interiordesignsuperm}{https://huggingface.co/stablediffusionapi/interiordesignsuperm}),

DVarch (\href{https://huggingface.co/stablediffusionapi/dvarch}{https://huggingface.co/stablediffusionapi/dvarch}).



\begin{figure}
    \centering
    \includegraphics[width=0.8\textwidth]{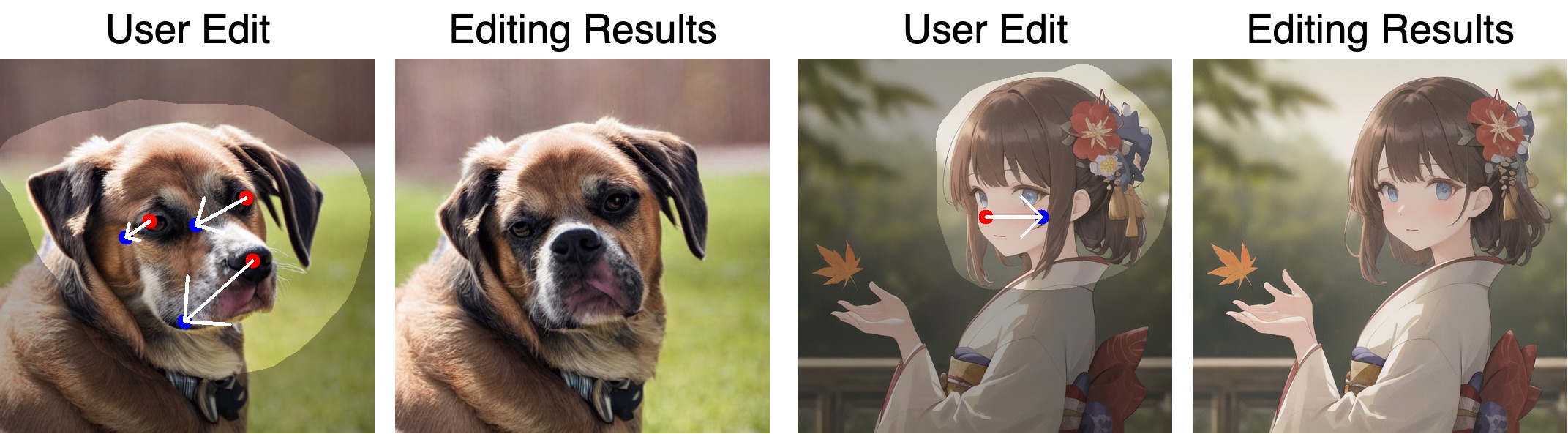}
    \vspace{-.2cm}
    \caption{Limitation of {\sc DragDiffusion}. Occasionally, some of the handle points cannot precisely reach the desired target.}
    \vspace{-.4cm}
    \label{fig:limitation}
\end{figure}

\section{More Details on Editing Diffusion-Generated Images}
Here we introduce more details about editing diffusion-generated images. Firstly, different from editing real images, we \textbf{do not} need to conduct LoRA fine-tuning before latent optimization. This is because the purpose of LoRA fine-tuning is to help better encode the features of the original image into the diffusion UNet. However, for diffusion-generated images, the image features are already well-encoded as the diffusion model itself can generate these images. In addition, during the latent optimization stage, we do not have to perform DDIM inversion as the diffusion latents are readily available from the generation process of the diffusion models.

Another details we need to attend to is the presence of classifier-free guidance (CFG) when editing generated images. As described in the main text, when editing real images, we turn off the CFG as it pose challenges to DDIM inversion. However, when editing generated images, we inevitably have to deal with CFG, as it is one of the key component in diffusion-based image generation. CFG introduces another forward propagation pass of the UNet during the denoising process with a negative text embedding from null prompt or negative prompt. This makes a difference during our latent optimization stage, as now we have two UNet feature maps (one from the forward propagation with positive text embedding and the other one from the negative text embedding) instead of only one. To deal with this, we concatenate these two feature maps along the channel dimension and then use the combined feature maps to supervise latent optimization. This simple strategy have been proven to be effective as shown in our empirical results.

\section{Execution Time}
Given a real image with the resolution of $512\times 512$, the execution time of different stages in {\sc DragDiffusion} on a A100 GPU is as follows: LoRA fine-tuning is around $25$ seconds, latent optimization is around $10$ to $30$ seconds depending on the magnitude of the drag-instruction, the final Latent-MasaCtrl guided denoising is negligible comparing to previous steps (about $1$ to $2$ seconds)

\section{Limitations}
\label{app:limitations}
As shown in Fig.~\ref{fig:limitation}, the limitation of our {\sc DragDiffusion} is that, occasionally, some of the handle points cannot precisely reach the desired target. This is potentially due to inaccurate point-tracking or difficulties in latent optimization when multiple pairs of handle and target points are given. We leave making the drag-based editing on diffusion models more robust and reliable as our future work.

\section{More Qualitative Results}
\label{app:more_qualitative}
To start with, we provide more qualitative comparisons between {\sc DragDiffusion} and {\sc DragGAN} in Fig.~\ref{fig:app_compare_draggan}. These results consistently showing that our approach demonstrate much better versatility than {\sc DragGAN}. 

\begin{figure}
    \centering
    \includegraphics[width=\textwidth]{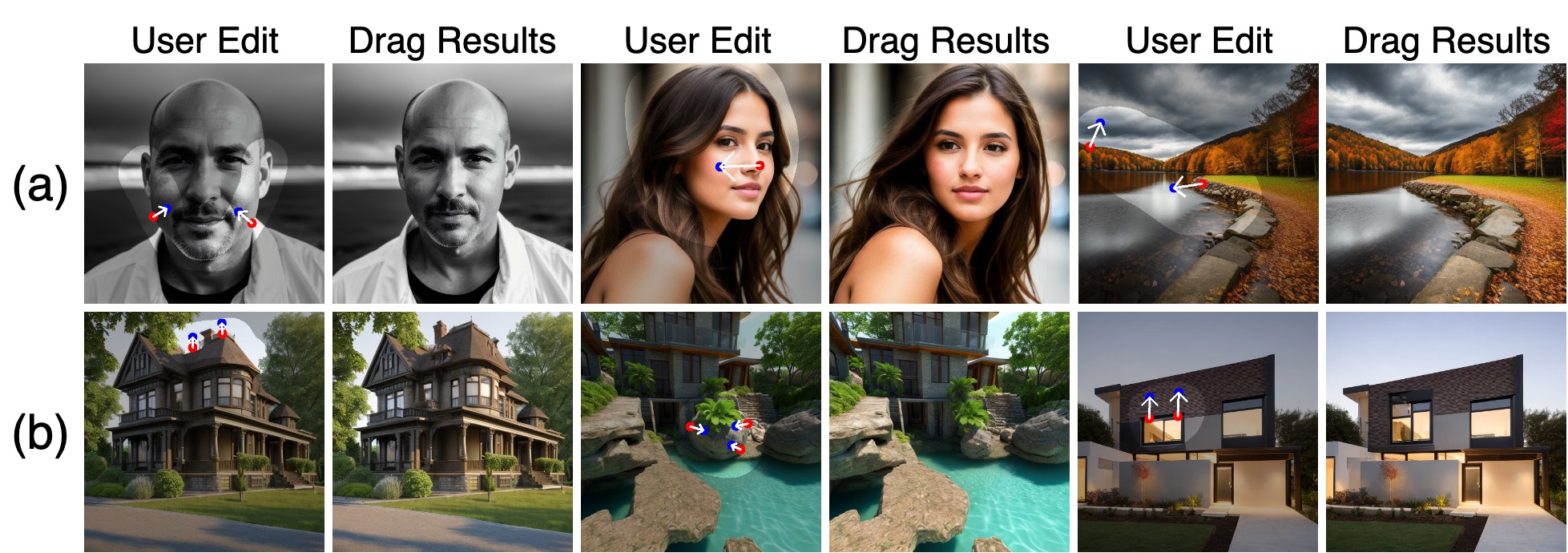}
    \caption{Editing results on diffusion-generated images with \textbf{(a)} Realistic Vision; \textbf{(b)} DVarch.}
    \label{fig:app_drag_generated}
\end{figure}

\begin{figure}
    \centering
    \includegraphics[width=\textwidth]{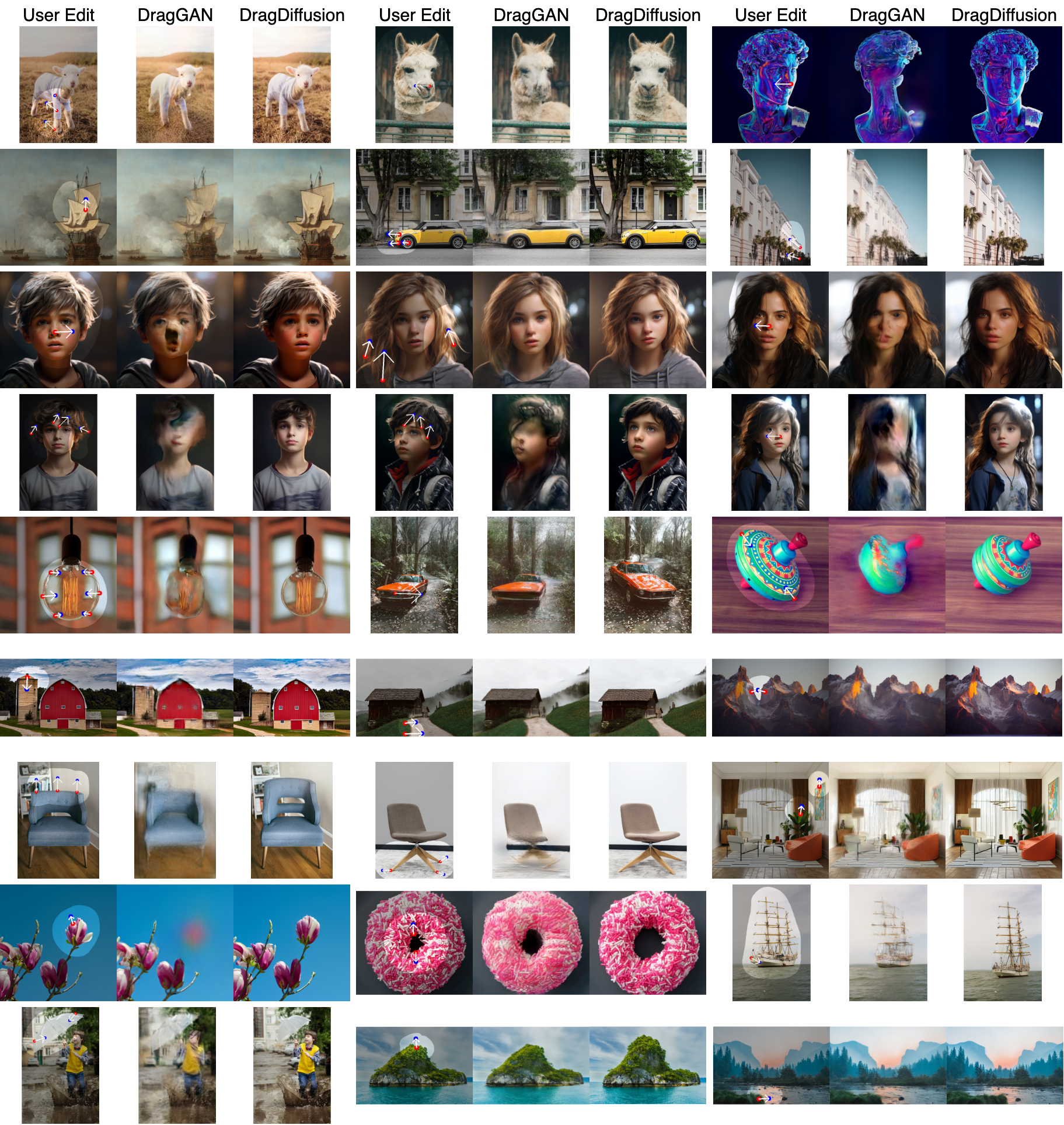}
    \caption{Additional comparisons between {\sc DragGAN} and {\sc DragDiffusion}. All images are from our {\sc DragBench} dataset. Both approaches are executed under the same drag-based editing instruction. \textbf{Zoom in to check the details.}}
    \label{fig:app_compare_draggan}
\end{figure}

Next, we demonstrate results of applying {\sc DragDiffusion} on images generated by two more fine-tuned variants of Stable-Diffusion-1.5, namely Realistic-Vision and DVarch. Results are shown in Fig.~\ref{fig:app_drag_generated}. These results along with the results in the main text corroborate the generality of our approach on different diffusion models.

\begin{figure}
    \centering
    \includegraphics[width=\textwidth]{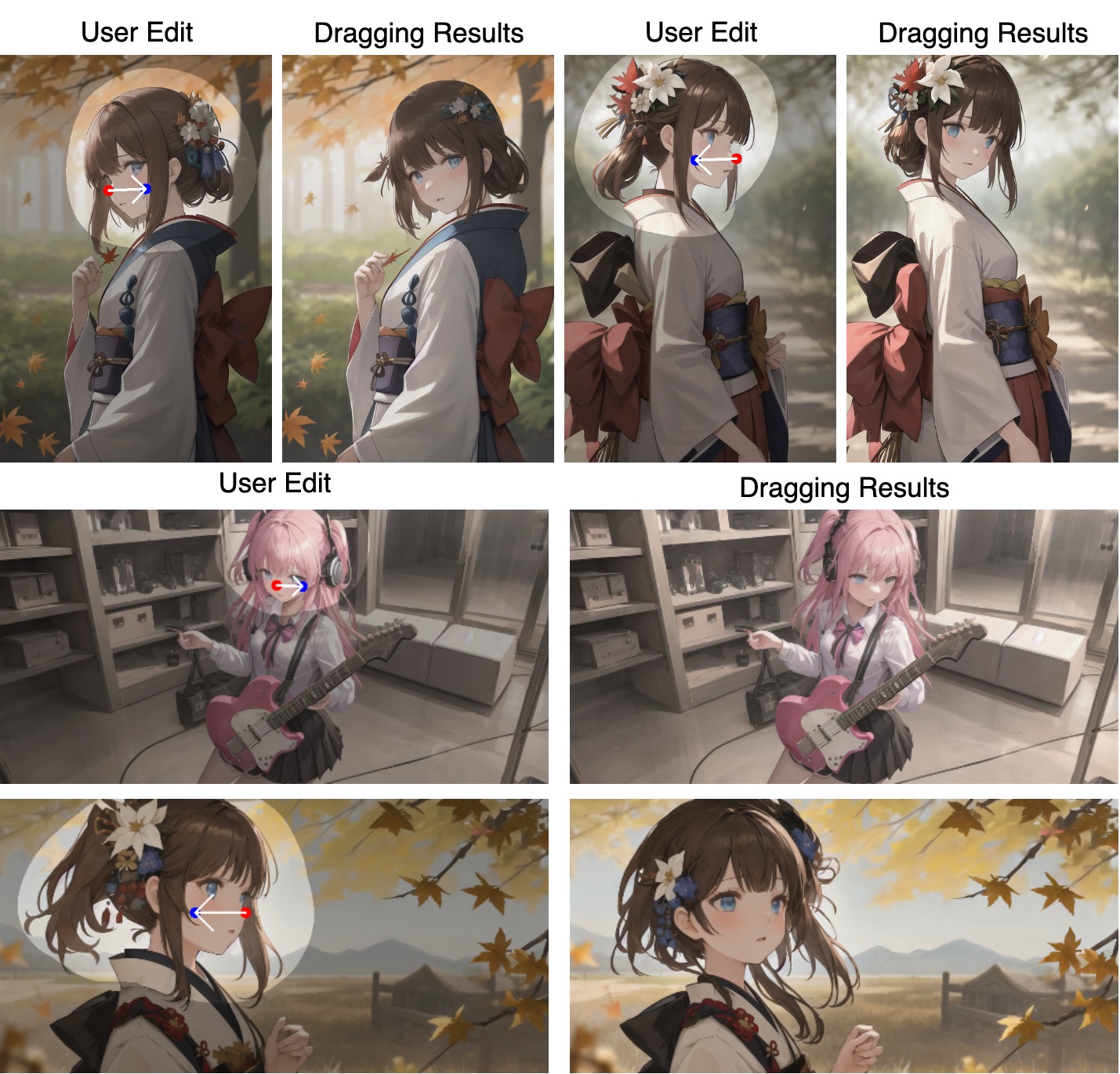}
    \caption{Editing results from {\sc DragDiffusion} beyond $512\times 512$ resolution. Results are produced by perform drag-based edits on images generated by Counterfeit-V2.5. The resolution of images in the first row are $768\times 512$, while the images in the second row are $512\times 1024$.}
    \label{fig:qual_more_resolution}
\end{figure}

Finally, we provide more results on generated images beyond the $512\times 512$ resolution as in the main text. These results are shown in Fig.~\ref{fig:qual_more_resolution}, which further demonstrate the versatility of {\sc DragDiffusion}.

\section{Visual Ablation on the Number of Identity-preserving fine-tuning steps}
\begin{figure}
    \centering
    \includegraphics[width=\textwidth]{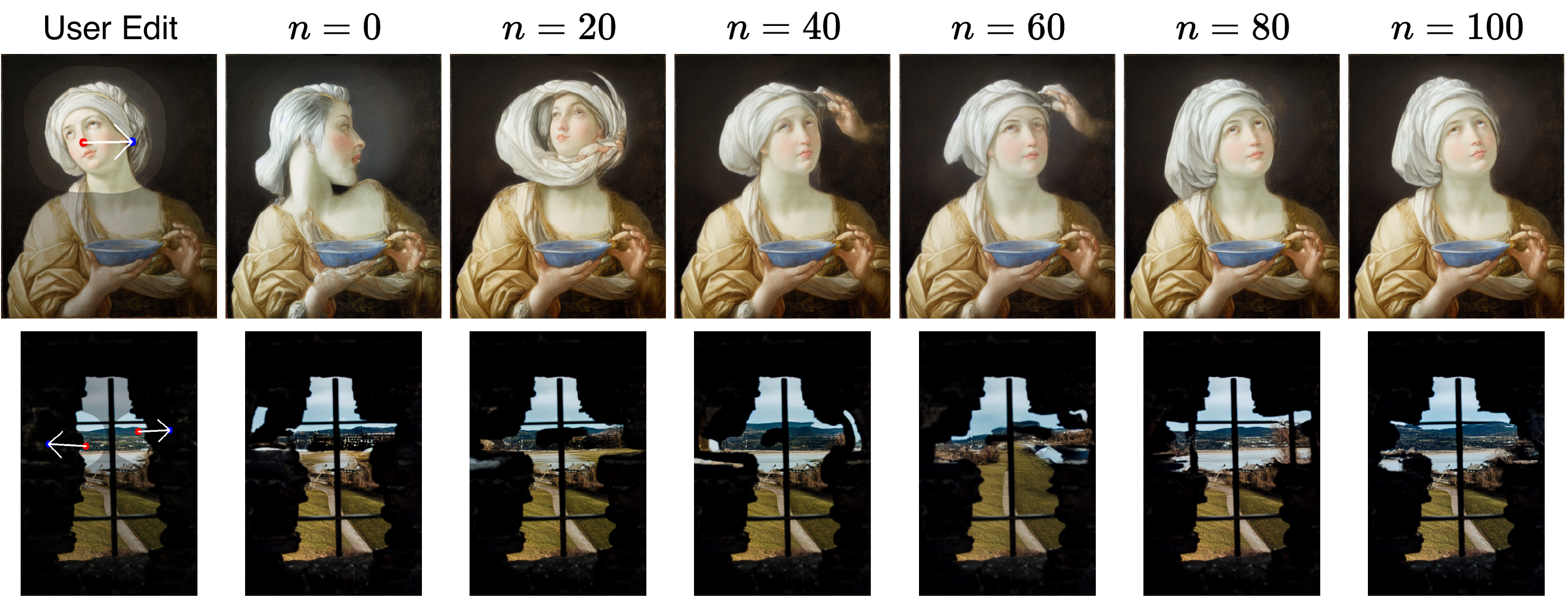}
    \caption{Visual ablation study on the number of \textbf{identity-preserving fine-tuning steps (denoted as $n$)}. \textbf{Zoom in to view details.} From the visualization, we see that setting $n<80$ can produce undesired artifacts in the dragging results (\eg, distorted faces and scenes, unexpected hands, \etc). On the other hands, $n\geq 80$ normally produces reasonable results without artifacts.}
    \label{fig:app_lora_steps}
\end{figure}

In the main paper, we ablate on the effect of the number of identity-preserving fine-tuning steps (denoted by $n$ in this section). We show through numerical experiments that $n \geq 80$ produce ideal results in Fig.~7~(b) of the main text. In this section, we provide visualization that corroborate with conclusions in the main text, showing setting $n \geq 80$ produces editing results that are free from artifacts such as distorted faces and scenes, unexpected hands, \etc. Results are shown in Fig.~\ref{fig:app_lora_steps}.

\section{Visual Ablation on the UNet Feature Maps}
\begin{figure}
    \centering
    \includegraphics[width=\textwidth]{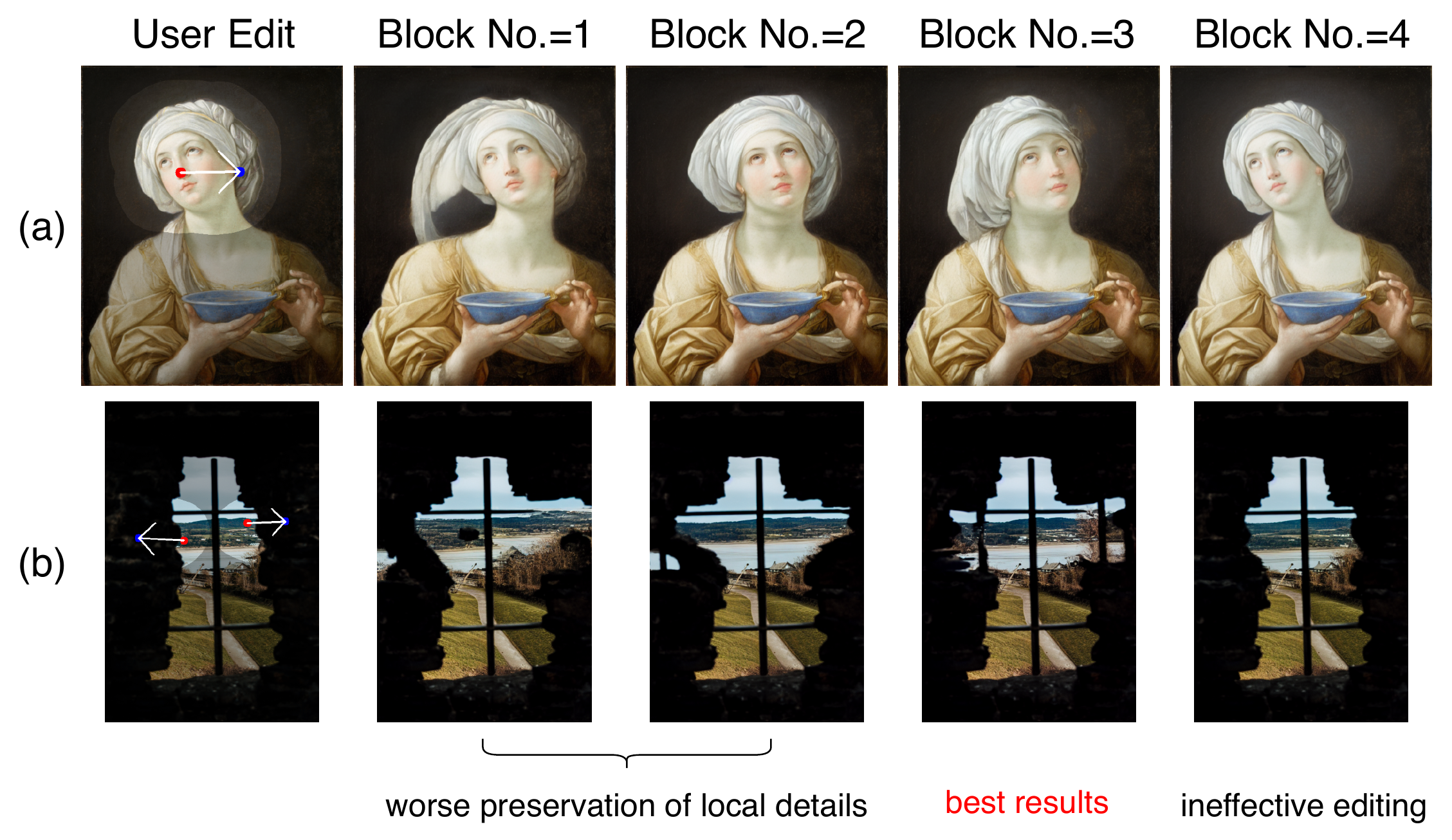}
    \caption{Visual ablation study on the block number of the UNet feature map. \textbf{Zoom in to view details.} As in the figure, using feature maps of the $2$-nd and $3$-rd blocks produce reasonable results. However, if observing more closely, we can see that using the $3$-rd block of features yields slightly better preservation of local details (\eg more reasonable headwrap in \textbf{(a)} and better details of buildings by the river in \textbf{(b)}).}
    \label{fig:app_unet_idx}
\end{figure}

In the main paper, we have studied the effect of using UNet feature maps produced by different blocks of UNet \textit{decoder} for our approach. In this section, we provide visualization that corroborates with conclusions in the main text. Results are shown in Fig.~\ref{fig:app_unet_idx}. According to the results, using the $1$-st block of feature maps will lead to unfavorable preservation of local details due to lack of fine-grained information. This corresponds to the low Image Fidelity (IF) and high Mean Distance (MD) as in main text Fig.~7~(c) when Block number is $1$.

On the other hand, as the $4$-th block of UNet feature maps only contains low-level information, the editing results is almost the same as the original real image, indicating ineffective editing. This corresponds to the high IF and high MD as in main text Fig.~7~(c) when Block number is $4$.

Finally, using the $2$-nd or the $3$-rd block of UNet feature maps can can yield reasonable editing. However, if observing more closely, we can see that using the $3$-rd block of features yields slightly better preservation of local details (\eg more reasonable headwrap in Fig.~\ref{fig:app_unet_idx}~(a) and better details of buildings by the river in the Fig.~\ref{fig:app_unet_idx}~(b)).
Correspondingly, in main text Fig.~7~(c), we also show using UNet feature maps output by the $3$-rd block can yield better results (lower MD and higher IF).

\clearpage

\section{Detailed Comparisons on {\sc DragBench} by Category}
In the main paper Fig.~8, we report the Mean Distance (MD) and Image Fidelity (IF) averaging over all samples in {\sc DragBench}. In this section, we provide detailed comparisons between {\sc DragGAN} and {\sc DragDiffusion} on each category in {\sc DragBench}. Comparisons in terms of IF (\ie, 1-LPIPS) and MD are given in Tab.~\ref{tab:detail_compare_if} and Tab.~\ref{tab:detail_compare_md}, respectively. According to our results, {\sc DragDiffusion} significantly outperforms {\sc DragGAN} in every categories of {\sc DragBench}.

\renewcommand{\arraystretch}{1.1}{
\setlength{\tabcolsep}{1mm}{
\begin{table*}[t]
\centering
\begin{tabular}{lcccccccccc}
\toprule
& art works & landscape & city & countryside & animals & head & upper body & full body & interior design & other objects  \\
\midrule
{\sc DragGAN} & 0.71 & 0.84 & 0.74 & 0.79 & 0.72 & 0.91 & 0.33 & 0.31 & 0.57 & 0.71 \\
\midrule
{\sc DragDiffusion} & 0.88 & 0.88 & 0.89 & 0.88 & 0.87 & 0.85 & 0.89 & 0.95 & 0.90 & 0.87 \\
\bottomrule
\end{tabular}
\vspace{-1mm}
\caption{\textbf{Comparisons of Image Fidelity (1-LPIPS) on {\sc DragBench} on each category ($\big\uparrow$).}}
\label{tab:detail_compare_if}
\end{table*}
}
}

\renewcommand{\arraystretch}{1.1}{
\setlength{\tabcolsep}{1mm}{
\begin{table*}[t]
\centering
\begin{tabular}{lcccccccccc}
\toprule
& art works & landscape & city & countryside & animals & head & upper body & full body & interior design & other objects  \\
\midrule
{\sc DragGAN} & 59.51 & 47.60 & 41.94 & 46.96 & 60.12 & 65.14 & 82.98 & 37.01 & 75.65 & 58.25 \\
\midrule
{\sc DragDiffusion} & 30.74 & 36.55 & 26.18 & 43.21 & 39.22 & 36.43 & 39.75 & 20.56 & 24.83 & 39.52 \\
\bottomrule
\end{tabular}
\vspace{-1mm}
\caption{\textbf{Comparisons of Mean Distance on {\sc DragBench} on each category ($\big\downarrow$).}}
\label{tab:detail_compare_md}
\end{table*}
}
}


\end{document}